\pdfoutput=1
\documentclass[11pt]{article}

\usepackage{EMNLP2023}

\usepackage{times}
\usepackage{latexsym}
\usepackage{siunitx}
\usepackage{comment}

\usepackage[T1]{fontenc}
\usepackage[utf8]{inputenc}

\usepackage{microtype}

\usepackage{inconsolata}

\title{Does ChatGPT have Theory of Mind?}

\usepackage{booktabs}
\usepackage{graphicx}
\usepackage{float}
\author{Bart Holterman \\
  Utrecht University \\\And
  Kees van Deemter \\
  Utrecht University\\}

\begin{document}
\maketitle
\begin{abstract}
Theory of Mind (ToM) is the ability to understand human thinking and decision-making, an ability that plays a crucial role in social interaction between people, including linguistic communication. This paper investigates to what extent recent Large Language Models in the ChatGPT tradition possess ToM. We posed six well-known problems that address biases in human reasoning and decision making to two versions of ChatGPT and we compared the results under a range of prompting strategies. While the results concerning ChatGPT-3 were somewhat inconclusive, ChatGPT-4 was shown to arrive at the correct answers more often than would be expected based on chance, although correct answers were often arrived at on the basis of false assumptions or invalid reasoning. % KvD Do you agree with this assessment, Bart? Are there other conclusions that are worth stating here? 
\end{abstract}

\section{Introduction}
%In less than 3 months, ChatGPT has taken the world by storm. 
ChatGPT is a large language model(LLM) chatbot based on the OpenAI ChatGPT-3 language model \cite{aydin2022openai}. A wealth of research articles are starting to chart ChatGPT's strengths and weaknesses \cite{aljanabi2023chatgpt}. %In a nutshell, ChatGPT the program produces seemingly intelligent writing \cite{van2023chatgpt}, but it also sometimes gives answers that are made up %(“hallucination effect")
%\cite{shen2023chatgpt} or factually wrong \cite{MarcusG}., 
% KvD Add a reference to Ziwei Ji et al, Survey of Hallucination in Natural Language Generation
%making it a questionable tool in areas such as medical writing\cite{biswas2023chatgpt}, where accuracy is of the utmost importance.  

    We add to these insights by studying ChatGPT's ability to comprehend human Theory of Mind (ToM). We 
    %using ToM principles such as the ones described by Daniel Kahneman in his book "Thinking, Fast and Slow"\cite{kahneman2011thinking}. 
    chart ChatGPT 3 and 4's ability to answer a range of well-known ToM problems, focussing on problems that address known human biases (henceforth, ``fallacies").
    %ChatGPT versions three and four are tested, and their results will be compared. 
    Since ChatGPT is sensitive to subtle differences in the prompt (e.g., \citet{moghaddam2023boosting}), we chart how ChatGPT responds to prompts with different levels of detail. 
    
    %This research is scientifically relevant to the field of AI because it gives insight into a relatively new and promising AI tool, for which there are many utilizations not yet scientifically studied. This thesis will research a specific utilization, namely the way ChatGPT handles prompts about human decision-making, which has not yet been extensively studied. The results of this research can elucidate possibilities and limitations for further studies using ChatGPT.
%
   % ToM covers a wide variety of different abilities, including assessments of other people's knowledge (e.g., as in the Sally-Anne problem \cite{kosinski2023theory}), and predictions of decisions that people are likely to take. %ChatGPT's ability to think about the Sally-Anne problem was investigated in (Add REF Rahimi and Honey).
    Since we wanted to cover a broad range of ToM abilities, we took a well-known overview of biases in human reasoning and decision making, namely \citet{kahneman2011thinking}, and selected a representative problem from each of its main chapters. Every prompt we gave to ChatGPT was based on a problem described in the book; ChatGPT's answers were compared with the correct answers. For example, suppose we ask, about a question $Q$, ``Which answer to $Q$ is most likely to be given by an arbitrary subject: answer $A$ or answer $B$?"; then if earlier experiments have shown overwhelmingly that more people chose $A$ than $B$, then the correct way for ChatGPT to answer our question is to say that $A$ is the most likely answer to $Q$. This means that, when subjects overwhelmingly tend to give answers to $Q$ that are wrong %(as in the Linda problem, for example, see section ...), 
    then the ChatGPT answer we count as {\em correct} must mark that false answer as the most likely one.
    
\section{Related Work}

% KvD I'm inclined to change the structure: first introduce ToM, then say ChatGPT's lack of world knowledge has been singled out as one of its main weaknesses. Then explan that ToM is a particular kind of world kowledge. I'm not sure this story should be told in this section or elsewhere.
%An often noted problem in that connection is that these systems lack general world knowledge (DO WE HAVE A REFERENCE FOR THIS?)
%— One kind of “world" knowledge that is particularly important for human interaction is Theory of Mind (ToM). ToM is known to be of particular importance for human communication (e.g. Herb Clark, Using Language, Cambridge Univ. Press 1996), for example when a speaker uses information in "common ground" to refer to an object (e.g., Clark & Marshall 1981; K. van Deemter, Computational Models of Referring, MIT Press 2016).

   % The literature is starting to offer an overview of the capabilities of ChatGPT \cite{biswas2023role,andrew2007scalable,zhang2021commentary}, including its limitations \cite{sobieszek2022playing,ji2023survey}. %Its tendency to generate outputs that are factually incorrect, unsupported by data, or otherwise purely invented by the model is called the ``hallucination effect". KvD DO WE NEED TO SAY THIS? IN THAT CASE, WE NEED TO SAY IT EARLIER AND REFER TO THE PAPERS WHERE NEURAL HALLUCINATION WAS FIRST NOTICED -- THAT'S WELL BEFOR THE ADVENT OF LLMs.

Theory of mind (ToM) is the ability to reason about mental states, such as beliefs, desires, and intentions, and to understand how mental states feature in everyday explanations and predictions of people's behavior \cite{apperly2012theory}. A variety of studies examine the presence and extent of ToM in large language models (LLM’s) \cite{kosinski2023theory, borji2023categorical, dou2023exploring, brunet2023conversational, moghaddam2023boosting}. These studies tend to employ false-belief tasks such as the “Sally-Anne Test”, a widely used test originally designed for assessing ToM capabilities in infants. %While this provides valuable insights into the system's ability to reason about beliefs, it only partially addresses the broader definition of ToM. %It falls short of comprehensively evaluating the system's understanding of other mental states including desires and intentions.
    This paper summarizes conducting an evaluation of ChatGPT’s understanding of human ``fallacies", %This entails an extension of the methodology to include reasoning about 
    looking at a range of mental states such as desires and intentions. Fallacies are a key aspect of ToM, because they show where humans behave in ways that deviate from what might be expected on purely ``rational" grounds.
    %With this addition, this study aims to provide a more nuanced understanding of ChatGPT’s ToM.

Our work bears some similarities to
    %Marcel Binz and Eric Schulz conducted a similar study
    \citet{binz2023using}'s study of ChatGPT's reasoning and decision making. %focused on understanding ChatGPT's cognitive psychology by 
    These authors, however, asked questions of the form ``what would you (ChatGPT) choose?". By contrast, our paper poses questions of the form, ``What would a third person be likely to choose", thus focussing on ChatGPT's understanding of human tendencies.

\section{Experiments}
Table 1 provides an overview of our approach to prompt engineering. To account for ChatGPT's stochastic nature, each prompt was presented nine times (n=9), in such a way that it can answer ’A’ one time and ’B’ another time the same prompt is posed. %Posing the prompt multiple times allows us to determine if there is a significant difference between the number of times ChatGPT provides answer ’A’ versus answer ’B’.
 %   \newline
    \begin{table}[H]
    \scalebox{0.8}{
    \begin{tabular}{@{}lllp{6cm}@{}}
     & Detailedness & Id & Type of prompt that was used\\ 
     \midrule
     & low & p1 & {A basic version of the experiment, with minimal details.} \\
    \midrule
        & medium & p2 & {Prompts are made more concrete, e.g., by giving people names.} \\
        %\addlinespace %\addlinespace %\addlinespace
        & & p3 & {More detail is added, making the scenario less abstract.} \\
    \midrule
        & high & p4 & {Further extraneous details are added.} \\
        %\addlinespace %\addlinespace %\addlinespace
        &  & p5 & {Like p4, but with a different choice of details.} \\
    \end{tabular}} 
    \caption{Our approach to prompt engineering}
    \end{table} 
    %
    %Through the analysis of the obtained results, we can draw conclusions about the answer that ChatGPT most commonly provides for a given prompt. Analysis will be done using a binomial test. Such a statistical test is important to analyze whether a correct answer is a result of mere chance or not. A binomial test resulting in a p-value of p < 0.05 indicates that the difference between the number of correct and incorrect answers is substantial enough to conclude that it is not a result of mere chance.

 %   After all answers are obtained, the significant answers will be compared to the correct answers as stated in the book ``Thinking, Fast and Slow" by Daniel Kahneman. The study will conclude with an evaluation of ChatGPT's ability to comprehend human decision-making. Additionally, the results per prompt will be observed to identify any common patterns, which can provide insight into how prompt detailedness affects ChatGPT's ability to answer Theory of Mind questions. In the upcoming section, each experiment will be explained in detail.
    
    \subsection{Experiments}
We briefly describe each of the six problem areas that featured in our experiments. The concrete versions of these problem areas that were used can be found in Appendix A.
   % As stated in the previous paragraph, six ToM problems were selected, each of which will be tested individually in an experiment. For each experiment, five prompts per experiment will be engineered with an increasing level of detail. Detailed explanations of each experiment will be presented in the upcoming sections. 
    \subsubsection{The Mental Shotgun}
    ``When we intend to complete one task, we involuntarily compute others at the same time, slowing us down" \cite{kahneman2011thinking}. Kahneman and colleagues investigated this hypothesis by having participants listen to spoken word pairs; participants were asked to press a button if and only if the pair rhymes. It turned out that, for word pairs that rhyme, morphological similarity speeds up the time lag, so ``vote-note" is recognized as rhyming more quickly than ``vote-goat" \cite{shotgun}. %Kahneman relates this to the mental shotgun effect, arguing that the subjects involuntarily computed how the words in the pair are written, slowing them down when the words are morphologically less similar. Our study will present this experiment to ChatGPT, posing the question for which word pair a human subject would press the button faster. In the prompts created (see Appendix A.1) in this study, the word pairs ``vote-goat" vs. ``vote-note" are used in p4. In p5 ``boat-coat" vs. ``boat-moat" is used. In p1,p2, and p3 no word examples are given. These prompts solely describe the final syllable of the word pairs. Through this experiment, we can infer whether ChatGPT has a comprehension of the mental shotgun principle. 
    Our study (see Appendix A.1) presented this experiment to ChatGPT, asking the system for which word pair a human subject would be likely to press the button faster. In prompts p1,...,p4, the same word pairs are presented as the ones used by Kahneman i.e.``vote-note" and ``vote-goat". In prompt p5, different word pairs are presented i.e. ``boat-coat" and ``boat-moat".
    
    \subsubsection{The Anchoring Effect}
    ``The anchoring effect occurs when people consider a particular value for an unknown quantity before estimating that quantity. What happens is one of the most reliable and robust results of experimental psychology: the estimates stay close to the number that people considered" \cite{kahneman2011thinking}. %The anchoring effect suggests that individuals can be subconsciously primed to make incorrect estimates due to previously heard quantities, called ``anchor points". Kahneman describes an experiment in which participants are asked if the famous painter Salvador Dali was younger or older than a certain age, $x$. He found that the participants would answer an age, close to the age $x$. To clarify, if $x=115$, participants would estimate a greater age, than if $x=65$. Kahneman assigns this to the anchoring effect, priming participants with the anchor, $x$. 
    Our experiment (Appendix A.2) is derived from the original one \cite{Anchoring}. In prompts p1-4 we used the same experiment as Kahneman, but we used different names and ages. In p5, an entirely different instance of the problem was created. %Specifically, a case in which the demand for a prison sentence is used as the anchor. Through this experiment we can infer whether ChatGPT has a comprehension of the cognitive bias, people have when presented with scenarios involving the anchoring effect.
    \subsubsection{The Linda Problem}
    The Linda problem %is an experiment created by Kahneman to test cognitive bias in decision-making. The consistency bias is described as a cognitive bias causing 
    addresses the human tendency to give undue weight to information that is in line with our pre-existing beliefs or expectations \cite{Linda}. %, even when such information is less likely to be true. In the Linda problem, 
    In earlier experiments, participants were presented with a scenario about a person called Linda, about whom they are told,``Linda is very active in equal rights movements". The subject is asked to choose the most likely option from a set of possibilities. One of the options involves a conjunction of two traits, such as ``Linda is a bank teller and a feminist", while another option only mentions just one of these traits, e.g., ``Linda is a bank teller". Perhaps surprisingly, participants tend to choose the option involving two traits, thus violating some elementary laws of probability. %Despite the fact that the option involving two traits is statistically less likely than the option involving one trait, the participants tended to choose the option involving two traits. However, this would be a conjunction fallacy, violating the logic of probability. Kahneman assigns this fallacy in decision-making to the consistency bias, explaining that our intuition favors what is plausible but improbable over what is implausible and probable\cite{kahneman2011thinking}. 
    In our experiment (see Appendix A.3), we modified the scenario so it does not involve the name Linda, nor the terms bank teller and feminist. %This is done because the Linda problem is a famous experiment and it is likely that there are instances of the exact Linda problem in ChatGPT's training data. %The scenario in our experiment (see Appendix A.3) describes a person called Robin (p1,...,p4), ChatGPT is then asked which option a third person will most likely choose, option A; Robin is a teacher and a woman, or option B; Robin is a teacher. The other example in our experiment (p5) describes a scenario of a man called Jan, ChatGPT is then asked to choose which option a third person will most likely choose, option A; Jan is a pilot and likes learning new languages or option B; Jan is a pilot. A noteworthy remark to make here is that the correct answer is the answer where the third person chooses the conjunction. This might appear counterintuitive, however, the study is about the comprehension of human ToM, which includes the fallacies humans tend to commit. Through this experiment we can infer whether ChatGPT has a comprehension of the cognitive bias, people have when presented with the Linda problem.
    \subsubsection{The Planning Fallacy}
    %The book describes the planning fallacy as how humans are often optimistic and confident of the best-case scenario and 
    Humans frequently “make decisions based on delusional optimism rather than on a rational weight of gains, losses, and probabilities,” \cite{kahneman2011thinking}; see also \citet{Optimism}. %The book states that humans often are surprised by unexpected costs, or missed deadlines, due to overconfidence. An example the book gives is when someone wants to remodel their kitchen, more often than not, the cost and duration of the remodeling are underestimated. 
    In our experiment (Appendix A.4), we used a kitchen remodeling scenario (p1-4) and a scenario of a project for a new subway line in Amsterdam (p5) as the prompts to pose to ChatGPT. %These examples of the planning fallacy are also described by Kahneman.
    %Through this experiment we can infer whether ChatGPT has a comprehension of the overconfidence principle, people tend to have when presented with scenarios potentially prone to a planning fallacy.
    
    \subsubsection{Relative Wealth}
%    In the book, Kahneman argues a point made by mathematician Daniel Bernoulli. Bernoulli theorizes that an absolute amount of money has only intrinsic value, saying ``A million dollars, is worth a million dollars, right?". Kahneman argues this theory with the example by saying
``Magically making a poor person’s portfolio worth a million dollars would be fabulous! Magically making a billionaire’s portfolio worth a million dollars would be agony!"\cite{kahneman2011thinking}. In other words, the subjective value, to a given individual, of that individual's wealth is relative to this individual's wealth in the recent past (see \cite{Wealth} for extensive discussion). %In our experiment (see Appendix A.5) we posed the scenario of two people with the same amount of money;  one of them gained money, while the other lost money. 
In versions p1-4 of our experiment (see Appendix A.5), the question posed to ChatGPT is whether a third person would think one of the two people was happier with an amount of money than the other, or if they are both equally happy. In version p5, financial wealth is replaced by physical fitness. In p1-4 we presented the same experiment as described in by Kahneman. In p5, an entirely different instance of the problem was created.
%We acknowledge that p5 is not about wealth in terms of money, however, the principle of relative happiness still applies. Through this experiment, we can infer whether ChatGPT has a comprehension of the prospect theory.

    \subsubsection{Loss aversion}
    Human loss aversion is extremely well supported (e.g., \citet{Loss}). The term describes how, ``when directly weighted against each other, losses loom larger than gains" \cite{kahneman2011thinking}.  This leads to a tendency to make choices that would be irrational when viewed in terms of expected value. %, as the prospect of a loss is perceived as more psychologically significant than the possibility of a gain. 
    In our experiment (see Appendix A.6), a scenario was created with a person presented with a bet with a positive expected value. According to Kahneman's theory of loss aversion, an individual would decline this bet. We used the same scenario as Kahneman describes, however we changed the amount of money. %Through this experiment, we can infer whether ChatGPT has a comprehension of the human tendency to be loss averse.     
    
\section{Results}
    GPT-3 outputs were obtained using ChatGPT version 3 with its last update on 24 March 2023. Outputs for GPT-4 were obtained using ChatGPT version 4 with its last update on 24 March 2023. %This is important due to the fast improvement of the model's performance, which can change drastically between updates. Although research shows and OpenAI openly admits \cite{OpenAI2022}, 
    %that ChatGPT can generate factually incorrect responses, 
    We opted to use ChatGPT's user interface which is also used by the majority of the public. Therefore, the parameter values used in the study are the default values provided in the OpenAI sample code\cite{motoki2023more}. An important parameter is the temperature, with a default value of 0.7\cite{motoki2023more}. %The temperature setting will be discussed in more detail in the discussion section. 
    A total of 45 outputs were obtained for each individual problem and each model (n=45). In order to examine the impact of varying prompts, five prompts were used per ToM principle. Both ChatGPT-3 and ChatGPT-4 were presented with each prompt. For each ChatGPT version, every prompt was posed nine times (n=9). The results of each experiment will be presented separately in this section. In total, each model was tested 270 times on its ability to comprehend ToM, using varying experiments and prompts. Out of the 270 total questions, ChatGPT-3 answered 147 correctly, a binomial test returns a p-value of $p>0.05$ which is in-significant . ChatGPT-4 answered 224 out of the 270 total questions correctly, a binomial test gives a p-value of $(p<0.0001)$, which is significant. To test the effect of the level of detail of the input prompts, five different input prompts were created per experiment. Figure 1 shows the average number of correct answers over all experiments, by model and prompt. ChatGPT-4 answered more questions correctly than ChatGPT-3,
    %, to test whether this difference in correct answers is significant, 
    as revealed by a two-proportion Z-Test ($p<0.0001$). Results per individual experiment are shown in Appendix B.
    %Conclusions will be drawn in the conclusion section. 
\begin{figure}[H]
    \centering   
    \includegraphics[width=8cm]{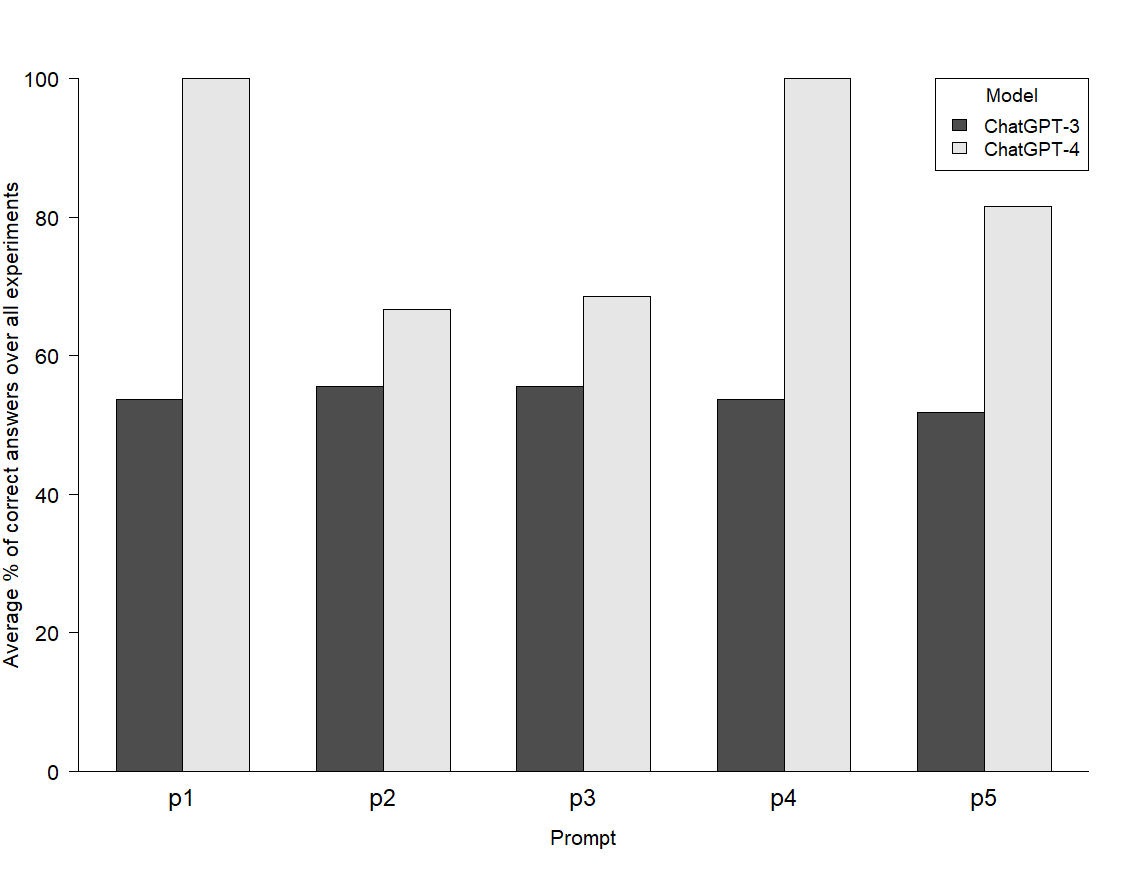}
    \caption{A bar plot showing the average number of correct answers over all experiments, by model and prompt. ChatGPT-3 has a similar correct answer average over all prompts, while ChatGPT-4 shows differences in the averages of correct answers over each prompt.}
    \label{bar}
\end{figure}
\raggedbottom
\section{Conclusion}    
    % \subsection{Academic embedding}
    % The findings of this paper are scientifically relevant to the field of AI because it gives an insight into a relatively new and promising AI tool, for which there are many utilizations not yet scientifically studied. This research studied a specific application of the model, namely, how does ChatGPT perform when posed with questions about human decision-making which is a utilization that has not been extensively studied yet. The results of this study can add to the existing body of research on ChatGPT and contribute to a deeper understanding of its capabilities and limitations. Additionally, this paper offers possible directions for further research.
    %\subsection{ChatGPT-3}
    This paper reports on an investigation into ChatGPT's knowledge and understanding of an important but complicated area of knowledge, namely the nature of human bias in reasoning and decision making.
    Summing up, ChatGPT-3 did not answer significantly more than half the total number of questions correctly ($p>0.05$), although it 
    %Therefore, we can not conclude that ChatGPT-3 performs well in the overall comprehension of ToM. However, observing the results for each individual experiment, it can be concluded that 
    performed somewhat better on the Linda problem, the planning fallacy and loss aversion.%This might be a result of increasing the detailedness that was not done systematically, as discussed in the limitations subsection.
   % \subsection{ChatGPT-4}
   %
    However, ChatGPT-4 performed much better, answering around 80\% of the total number of questions correctly.
    %Therefore we can conclude that ChatGPT-4 performs well in the overall comprehension of ToM. When observing the individual experiments, it can be concluded that 
    The system performs particularly well on the mental shotgun, the Linda problem, the Anchoring effect, and the Planning Fallacy. For unknown reasons that may be explored in future research, ChatGPT-4 performed better with the low and high-detailed prompts compared to the mid-detailed prompts. 
   Thus, our results could be taken to suggest that ChatGPT does not, but ChatGPT-4 does, have a certain amount of comprehension of human biases in reasoning and decision making, a key facet of ToM. Broadly speaking, these results are in line with earlier findings\cite{borji2023categorical}\cite{brunet2023conversational}, 
   which ascribed a certain amount of ToM to ChatGPT, particularly regarding false-belief tasks (as discussed in section 2). But although these results hold promise for a future in which Artificial Intelligence might reach a genuine understanding of people, we believe that our results, and earlier research on ChatGPT's ToM as well, \cite{kosinski2023theory}\cite{moghaddam2023boosting}, might flatter ChatGPT's actual abilities.

One reason is that systems can be ``right for the wrong reasons" \cite{mccoy2019right}. So although this was not the original focus of our work, we decided to examine the {\em explanations} (Appendix C) ChatGPT gave for its decisions, in order to test the quality of the systems' reasoning. Because the reasoning is not always clear (and occasionally completely {\em un}clear, as in the case of the Anchoring Effect, Appendix C2.1 and C2.2), such an analysis is not easy to perform, but in some cases, we detected flaws in the argumentation even in cases where the correct answer was chosen.
   
   For example, in the answers to the Mental Shotgun, ChatGPT3 repeatedly explains its answer offering an incorrect assessment of whether two words rhyme. (Appendix C1.1 and C1.2). An interesting situation was encountered in the analysis of the Loss Aversion problem. In its first response (C6.1), the system provides the correct answer. However, the model incorrectly claims that the expected value of the bet is negative. In the second instance, the model provides an answer that does not take into account the human tendency to be loss averse and is therefore recorded as incorrect. Nonetheless, in contrast to the first instance, the system correctly calculates the expected value.

  These observations lead on to a worry, with implications for all experimental evaluations of ChatGPT, about what may be called ``parrotting". Despite our best efforts, it is conceivable that, when faced with a ToM problem, ChatGPT bases its answers on discussions of that same problem, or a closely related problem, in its training material. For instance, if experimehts demonstrated that ChatGPT can prove Pythagoras's theorem, we wouldn't jump to the conclusion that ChatGPT can invent mathematical proofs in geometry; the system's proof may simply have been found in the texts on which ChatGPT was trained. 
   
The study reported in this paper mitigated this problem by changing the original problems in all the prompts, and additionally varying the prompts very considerably, looking at 5 different versions of each problem. Except perhaps in the case of the Mental Shotgun (where the nature of the problem make it challenging to vary the topic), especially version p5 of each problem was always completely different from versions of the problem that we were able to find in existing experiments. Yet Fig. 1 shows that performance on p5 was generally in line with performance on p1-4, suggesting that similarity to training material did not influence performance unduly. In future work, these issues should be investigated further, for example by perturbing the information in the prompts more systematically (e.g., by judiciously adding negations, cf. \citet{niven2019probing}) and probing how these perturbations affect ChatGPT's responses.
   
   Far from considering these matters settled, we believe that deep questions are lurking in the background: How ``novel" does a problem, in any area, have to be so we can count it as a fair test of ChatGPT's understanding of that area? Suppose, for example, a famous theme like Loss Aversion was discussed in the literature so extensively, in such detail and with so many examples, that a good proportion of ToM problem could be solved by learning from these discussions (e.g., using a state-of-the-art Natural Language Inference system (e.g., \citet{schick2020exploiting}). Under these assumptions, the system would tend to perform above chance level on any reasonable test battery for Loss Aversion. If this happened, would this mean that ChatGPT ``understands" Loss Aversion? Returning to the title of the paper, what does it mean to ``have" Theory of Mind? These questions, which call to mind Searle's Chinese Room experiment \cite{searle}, are all the more pertinent given that psychologists have learned to be cautious to distinguish between having a capability and actually using that capability. These considerations are all the more relevant because some of the most cogent experimental results underpinning the distinction between having and using a capabilty come precisely from the study of human Theory of Mind, where researchers have found strong limitations of the circumstances under which people use ToM (see e.g. \cite{keysar2003limits} for elaboration and experiments).

\newpage
\bibliography{emnlp2023}

\begin{thebibliography}{22}
\expandafter\ifx\csname natexlab\endcsname\relax\def\natexlab#1{#1}\fi

\bibitem[{Aljanabi et~al.(2023)Aljanabi, Ghazi, Ali, Abed
  et~al.}]{aljanabi2023chatgpt}
Mohammad Aljanabi, Mohanad Ghazi, Ahmed~Hussein Ali, Saad~Abas Abed, et~al.
  2023.
\newblock Chatgpt: Open possibilities.
\newblock \emph{Iraqi Journal For Computer Science and Mathematics},
  4(1):62--64.

\bibitem[{Apperly(2012)}]{apperly2012theory}
Ian~A Apperly. 2012.
\newblock What is “theory of mind”? concepts, cognitive processes and
  individual differences.
\newblock \emph{Quarterly Journal of Experimental Psychology}, 65(5):825--839.

\bibitem[{Ayd{\i}n and Karaarslan(2022)}]{aydin2022openai}
{\"O}mer Ayd{\i}n and Enis Karaarslan. 2022.
\newblock Openai chatgpt generated literature review: Digital twin in
  healthcare.
\newblock \emph{Available at SSRN 4308687}.

\bibitem[{Binz and Schulz(2023)}]{binz2023using}
Marcel Binz and Eric Schulz. 2023.
\newblock Using cognitive psychology to understand gpt-3.
\newblock \emph{Proceedings of the National Academy of Sciences},
  120(6):e2218523120.

\bibitem[{Borji(2023)}]{borji2023categorical}
Ali Borji. 2023.
\newblock A categorical archive of chatgpt failures.
\newblock \emph{arXiv preprint arXiv:2302.03494}.

\bibitem[{Brunet-Gouet et~al.(2023)Brunet-Gouet, Vidal, and
  Roux}]{brunet2023conversational}
Eric Brunet-Gouet, Nathan Vidal, and Paul Roux. 2023.
\newblock Do conversational agents have a theory of mind? a single case study
  of chatgpt with the hinting, false beliefs and false photographs, and strange
  stories paradigms.

\bibitem[{Coombs et~al.(1970)Coombs, Dawes, and Tversky}]{Wealth}
C.H. Coombs, R.M. Dawes, and A.~Tversky. 1970.
\newblock \emph{Mathematical Psychology: An Elementary Introduction}.
\newblock Prentice-Hall, New York.

\bibitem[{Dou(2023)}]{dou2023exploring}
Zenan Dou. 2023.
\newblock Exploring gpt-3 model's capability in passing the sally-anne test a
  preliminary study in two languages.

\bibitem[{Jacowitz and Kahneman(1995)}]{Anchoring}
K.E. Jacowitz and D.~Kahneman. 1995.
\newblock Measures of anchoring in estimation tasks.
\newblock \emph{Personality and Social Psychology Bulletin}, 21:1161--66.

\bibitem[{Kahneman and Lovallo(2003)}]{Optimism}
D.~Kahneman and D.~Lovallo. 2003.
\newblock Delusions of success: how optimism undermines executives' decisions.
\newblock \emph{Harvard Business Review}, 81:56--63.

\bibitem[{Kahneman(2011)}]{kahneman2011thinking}
Daniel Kahneman. 2011.
\newblock \href
  {https://www.amazon.de/Thinking-Fast-Slow-Daniel-Kahneman/dp/0374275637/ref=wl_it_dp_o_pdT1_nS_nC?ie=UTF8&colid=151193SNGKJT9&coliid=I3OCESLZCVDFL7}
  {\emph{Thinking, fast and slow}}.
\newblock Farrar, Straus and Giroux, New York.

\bibitem[{Keysar et~al.(2003)Keysar, Lin, and Barr}]{keysar2003limits}
Boaz Keysar, Shuhong Lin, and Dale~J Barr. 2003.
\newblock Limits on theory of mind use in adults.
\newblock \emph{Cognition}, 89(1):25--41.

\bibitem[{Kosinski(2023)}]{kosinski2023theory}
Michal Kosinski. 2023.
\newblock Theory of mind may have spontaneously emerged in large language
  models.
\newblock \emph{arXiv preprint arXiv:2302.02083}.

\bibitem[{McCoy et~al.(2019)McCoy, Pavlick, and Linzen}]{mccoy2019right}
R~Thomas McCoy, Ellie Pavlick, and Tal Linzen. 2019.
\newblock Right for the wrong reasons: Diagnosing syntactic heuristics in
  natural language inference.
\newblock \emph{arXiv preprint arXiv:1902.01007}.

\bibitem[{Moghaddam and Honey(2023)}]{moghaddam2023boosting}
Shima~Rahimi Moghaddam and Christopher~J Honey. 2023.
\newblock Boosting theory-of-mind performance in large language models via
  prompting.
\newblock \emph{arXiv preprint arXiv:2304.11490}.

\bibitem[{Motoki et~al.(2023)Motoki, Pinho~Neto, and
  Rodrigues}]{motoki2023more}
Fabio Motoki, Valdemar Pinho~Neto, and Victor Rodrigues. 2023.
\newblock More human than human: Measuring chatgpt political bias.
\newblock \emph{Available at SSRN 4372349}.

\bibitem[{Niven and Kao(2019)}]{niven2019probing}
Timothy Niven and Hung-Yu Kao. 2019.
\newblock Probing neural network comprehension of natural language arguments.
\newblock \emph{arXiv preprint arXiv:1907.07355}.

\bibitem[{Novemsky and Kahneman(2005)}]{Loss}
N.~Novemsky and D.~Kahneman. 2005.
\newblock The boundaries of loss aversion.
\newblock \emph{Journal of Marketing Research}, 42:119--28.

\bibitem[{Schick and Sch{\"u}tze(2020)}]{schick2020exploiting}
Timo Schick and Hinrich Sch{\"u}tze. 2020.
\newblock Exploiting cloze questions for few shot text classification and
  natural language inference.
\newblock \emph{arXiv preprint arXiv:2001.07676}.

\bibitem[{Searle(1980)}]{searle}
J.~Searle. 1980.
\newblock Minds, brains and programs.
\newblock \emph{Behavioral and Brain Sciences}, 3:417--57.

\bibitem[{Seidenberg and Kahneman(1979)}]{shotgun}
M.S. Seidenberg and D.~Kahneman. 1979.
\newblock Orthographic effects on rhyme monitoring.
\newblock \emph{Journal of Experimental Psychology: Human Learning and Memory},
  5:546--54.

\bibitem[{Tversky and Kahneman(1983)}]{Linda}
A.~Tversky and D.~Kahneman. 1983.
\newblock Extensional versus intuitive reasoning: the conjunction fallacy in
  probability judgment.
\newblock \emph{Psychological Review}, 90:293--315.

\end{thebibliography}
\bibliographystyle{acl_natbib}
\newpage
\appendix
\newpage
\section{Appendix: Prompts per ToM problem}
\label{appendix:A}

\subsection{Prompts for the Mental Shotgun problem} 
\begin{table}[h]
    \scalebox{0.85}{
    \begin{tabular}{l@{}l p{.3cm} p{6.3cm} }
            & Detail & Id & Prompt \\
    \midrule
        & low & p1 & {What will a person recognize as being a rhyme faster? A: A word that ends with “oat” and a word that ends with “ote” or B: two words that end in “ote”.} \\
    \midrule
        & medium & p2 & {A person is being asked to listen to two pairs of words. Option A: a word that ends with “oat” and a word that ends with “ote” or option B: two words that end with “ote”. Which pair will the person recognize faster as rhyming?} \\
        \addlinespace \addlinespace \addlinespace
        & & p3 & {Jan is asked to listen to two pairs of words. While listening to the words Jan has to press a button when he thinks the pair of words rhyme. Pair A: consists of one word that ends with “ote” and one word that ends with “oat”. Pair B: consists of two words that both end in “ote”. For which pair will Jan press the button faster?} \\
    \midrule
        & high & p4 & {Jan is asked to listen to pairs of words. While listening to the words Jan must press a button when he thinks the pair of words rhyme. Pair A: consists of the words “note” and “goat”. Pair B: consists of the words “note” and “vote”. For which pair will Jan press the button faster?} \\
        \addlinespace \addlinespace \addlinespace
        &  & p5 & {Jan is asked to listen to pairs of words. While listening to the words Jan must press a button when he thinks the pair of words rhyme. Pair A: consists of the words, “boat” and “mote”. Pair B: consists of the words, “boat” and “coat”. For which pair will Jan press the button faster?} \\
    \end{tabular}}
\end{table}
\newpage
\subsection{Prompts for the Anchoring effect problem} 
\begin{table}[h]
    \scalebox{0.71}{
    \begin{tabular}{l@{}l p{.3cm} p{8.1cm} }
            & Detail & Id & Prompt \\
    \midrule
        & low & p1 & {One person is asked if van Gogh died before or after he was age 105. Another person is asked if van Gogh died before or after he was age 55. They are both asked to guess van Gogh’s age. Which option is most likely? Option A: The first person guesses a higher number than the second person Option B: The second person guesses a higher number than the first person.} \\
    \midrule
        & medium & p2 & {Two persons are participating in a questionnaire, the first person is asked if Vincent van Gogh died before or after he was age 105. The second person is asked if Vincent van Gogh died before or after he was age 55. The next question in the questionnaire is to guess the age at which Vincent van Gogh died. Which option is most likely? Option A: The first person guesses a higher number than the second person Option B: The second person guesses a higher number than the first person.} \\
        \addlinespace \addlinespace \addlinespace
        & & p3 & {Two persons are participating in a questionnaire, the first person, called John, is asked if Vincent van Gogh died before or after he was age 105. The second person, called Rick, is asked if Vincent van Gogh died before or after he was age 55. The next question in the questionnaire is to guess the age at which Vincent van Gogh died. Which option is most likely? Option A: John guesses a higher number than Rick Option B: Rick guesses a higher number than John.} \\
    \midrule
        & high & p4 & {Two persons are participating in a questionnaire, the first person’s name is John, John is a 27-year-old Graphic designer from Pittsburgh, he is asked if Vincent van Gogh died before or after he was age 105. The second person’s name is Rick, Rick is a 45-year-old lawyer from Seattle, he is asked if Vincent van Gogh died before or after he was age 55. The next question in the questionnaire is to guess the age at which Vincent van Gogh died. Which option is most likely? Option A: John guesses a higher number than Rick Option B: Rick guesses a higher number than John.} \\
        \addlinespace \addlinespace \addlinespace
        &  & p5 & {There are two criminals, their names are Paul and John. They have each individually committed tax fraud with which they have evaded 3 million dollars. For a crime like this, a normal sentence would be 5 years in prison. Paul and John have two different prosecutors. Paul’s prosecutor demands 7 years prison. John’s prosecutor demands 12 years in prison. Note, they have done the same crime. They both have the same judge. Which option has the highest probability? Option A:  One of them will get a higher sentence. Option B: Paul and John will get the same sentence.} \\
    \end{tabular}}    
\end{table}
\newpage
\subsection{Prompts for the Linda Problem} 
\begin{table}[h]
\hspace*{-1cm}
    \scalebox{0.8}{
    \begin{tabular}{l@{}l p{.3cm} p{8.42cm} }
            & Detail & Id & Prompt \\
    \midrule
        & low & p1 & {Robin is a person, who loves kids. A third person does not know Robin's employment or gender. Option A: Robin is a teacher and a woman or Option B: Robin is a teacher. Which one of these options is more likely to be chosen by the third person?} \\
    \midrule
        & medium & p2 & {Robin is a person who loves kids, likes to play board games, and is very active in equal rights movements. Furthermore, a third person has no knowledge of Robin's employment or gender. Option A: Robin is a teacher and a woman or Option B: Robin is a teacher. Which one of these options is more likely to be chosen by the third person?} \\
        \addlinespace \addlinespace \addlinespace
        & & p3 & {Robin is a person from the United Kingdom who loves kids, Robin is smart and likes sharing knowledge with others. Robin likes to play board games every Tuesday at a board game club. Additionally, Robin is very active in equal rights movements going to almost every protest in the nation (United Kingdom). Furthermore, Rick is a third person who has no knowledge of Robin's employment or gender. Option A: Robin is a teacher and a woman or Option B: Robin is a teacher. Which one of these options is more likely to be chosen by Rick?} \\
    \midrule
        & high & p4 & {Robin is a person from the United Kingdom who has a deep love for children and enjoys spending time with them. In their free time, they volunteer at a local children’s charity, where they organize and run educational activities for underprivileged children. Robin is smart and likes sharing knowledge with others. Robin likes to play board games every Tuesday at a board game club. Additionally, Robin is very active in equal rights movements going to almost every protest in the nation (United Kingdom). Furthermore, Rick is a third person who has no knowledge of Robin's employment or gender. Option A: Robin is a teacher and a woman or Option B: Robin is a teacher. Which one of these options is more likely to be chosen by Rick?} \\
        \addlinespace \addlinespace \addlinespace
        &  & p5 & {Jan is a man that has been all over the world, he has lived in 4 different countries, and visited 50 more countries. All this traveling is needed for his work. Furthermore, Jan has a wife and kids and loves to do crossword puzzles. Because Jan has lived in so many countries, he can speak 5 languages fluently. Robert doesn't know Jan’s employment. Option A: Jan is a pilot and likes learning new languages. Option B: Jan is a pilot. Which one of these options is more likely to be chosen by Robert?} \\
    \end{tabular}}   
\end{table}
\newpage
\subsection{Prompts for the Planning Fallacy} 
\begin{table}[h]
    \scalebox{0.85}{
    \begin{tabular}{l@{}l p{.3cm} p{6cm} }
            & Detail & Id & Prompt \\
    \midrule
        & low & p1 & {Someone wants to remodel their kitchen, he estimates it will be around a certain amount of money. Option A: the actual price will be around that amount of money. Option B: The actual amount of money will be higher. Which option is more likely?} \\
    \midrule
        & medium & p2 & {Rick is an average American, he wants to remodel his kitchen, Rick estimates the price he thinks the remodeling will cost. Option A: the actual price will be around that price. Option B: The actual price will be higher. Which option is more likely?} \\
        \addlinespace \addlinespace \addlinespace
        & & p3 & {Rick is an average American, he wants to remodel his kitchen by putting in a new counter and refrigerator, Rick estimates the price he thinks the remodeling will cost. Option A: the actual price will be around that price. Option B: The actual price will be higher. Which option is more likely?} \\
    \midrule
        & high & p4 & {Rick, an average American homeowner, has plans to renovate his kitchen by installing a new countertop and refrigerator. After conducting some research, he comes up with an estimated budget for the remodeling project. Option A: the actual price will be around that price. Option B: The actual price will be higher. Which option is more likely?} \\
        \addlinespace \addlinespace \addlinespace
        &  & p5 & {The city of Amsterdam wants to build a new metro line that goes from east to west Amsterdam it is estimated to cost 20 million euros and will be done in 2025. Option A: the actual price will be around that price. Option B: The actual price will be higher. Which option is more likely?} \\
    \end{tabular}}   
\end{table}
\newpage
\subsection{Prompts for the Relative Wealth problem} 
\begin{table}[h]
    \scalebox{0.75}{
    \begin{tabular}{l@{}l p{0.3cm} p{7.5cm} }
            & Detail & Id & Prompt \\
    \midrule
        & low & p1 & {Two men have the same amount of money. The first man lost money while the second man gained money, but now they have the same amount. Option A: A third person will likely think one of the two men is happier with his amount of money. Option B: A third person will likely think both men are equally happy with their amount of money} \\
    \midrule
        & medium & p2 & {Yesterday Rick gained 4 million dollars and Emma lost 4 million dollars, today they both have 5 million. A third person is asked who is happier with the amount of money they have, what will he likely say? Option A: one of the two is happier with the amount of money they have. Option B: They are both equally happy with the amount of money they have.} \\
        \addlinespace \addlinespace \addlinespace
        & & p3 & {Rick as of yesterday had 1 million dollars, Emma on the other hand had 9 million dollars as of yesterday, due to volatility in the market they both have the same amount of money, namely 5 million dollars. Tim hears about this and is asked who is happier with the amount of money they have. What will Tim likely say? A: One of the two is happier with the amount of money they have. Option B: They are both equally happy with the amount of money they have} \\
    \midrule
        & high & p4 & {Rick as of yesterday had 1 million dollars, Emma on the other hand had 9 million dollars as of yesterday. Emma lost 4 million dollars in a day while Rick gained 4 million dollars due to volatility in the market. This means they both have the same amount of money, namely 5 million dollars. Tim hears about this and is asked who is happier with the amount of money they have. Which option will Tim likely choose? A: one of the two is happier with the amount of money they have. Option B: They are both equally happy with the amount of money they have.} \\
        \addlinespace \addlinespace \addlinespace
        &  & p5 & {Devon and Marc regularly go to the gym to get stronger. They both bench press 80kg as a one rep max as of today. One month ago, Devon bench pressed 70kg while Marc bench pressed 90kg a month ago. Tim is a friend of theirs and is asked who is likely to be happier with their one rep max of 80kg today. Option A: One of them is happier with their one rep max of 80kg. Option B: They are both equally happy with their one rep max of 80kg.} \\
    \end{tabular}}   
\end{table}
\newpage
\subsection{Prompts for the Loss Aversion problem} 
\begin{table}[h]
    \scalebox{0.85}{
    \begin{tabular}{l@{}l p{.3cm} p{6cm} }
            & Details & Id & Prompt \\
    \midrule
        & low & p1 & {A person is presented with a choice 50\% chance to lose \$100 and a 50\% chance to win \$125. What is more likely? Option A: They take the bet. Option B: They do not take the bet.} \\
    \midrule
        & medium & p2 & {An average American is presented with a bet. The bet is a 50\% chance to lose \$100 and a 50\% chance to win \$125. Option A: They take the bet. Option B: They do not take the bet. Which option is most likely to be chosen by the average person?} \\
        \addlinespace \addlinespace \addlinespace
        & & p3 & {An average American is presented with a bet, the bet involves flipping a coin, if it comes up heads lose \$100, but if it comes up tails win \$125. Option A: The average American takes the bet. Option B: The average American does not take the bet. Which option is most likely to be chosen by the average American?} \\
    \midrule
        & high & p4 & {Julia is a 42-year-old accountant from New York City. One day, her friend offers her a bet. The bet involves flipping a coin, where if it comes up heads, she will lose \$100, but if it comes up tails, she will win \$125. She is given the option to take the bet or not. Option A: Julia takes the bet. Option B: Julia does not take the bet. Which option is most likely to be chosen by Julia?} \\
        \addlinespace \addlinespace \addlinespace
        &  & p5 & {Tom is a 28-year-old college student from Los Angeles. One day, his roommate offers him a bet. The bet involves flipping a coin, where if it comes up heads, he will lose \$200, but if it comes up tails, he will win \$250. He is given the option to take the bet or not. Option A: Tom takes the bet. Option B: Tom does not take the bet. Which option is most likely to be chosen by Tom?} \\
    \end{tabular}} 
\end{table}

\section{Appendix: Results per prompt, per ToM problem}
\label{appendix:B}

\subsection{Results for the Mental Shotgun problem} 
\begin{table}[H]
\scalebox{0.8}{
\begin{tabular}{@{}lllll@{}}
\toprule
 & Prompt & {Correct (\%)} & {\textit{p}-value} & {Result} \\ \midrule
 ChatGPT-3 & p1 & 56 & 1.0000 & Inconclusive \\
                  & p2 & 0 & 0.0039 & Wrong \\
                  & p3 & 0 & 0.0039 & Wrong \\
                  & p4 & 0 & 0.0039 & Wrong \\ \cmidrule{2-5}
                  & p5 & 11 & 0.0391 & Wrong \\
                  \addlinespace
    \midrule
    ChatGPT-4 & p1 & 100 & 0.0039 & Right \\
                  & p2 & 100 & 0.0039 & Right \\
                  & p3 & 100 & 0.0039 & Right \\
                  & p4 & 100 & 0.0039 & Right \\
                  \cmidrule{2-5}
                  & p5 & 100 & 0.0039 & Right \\
    \bottomrule
\end{tabular}}
\caption{Shows, for each model, for each prompt, the percentage of correct answers, the p-value from a binomial test, and the result that can be concluded given the p-value.}\label{tab:final_expression}
\end{table}

\subsection{Results for the Anchoring effect problem} 
\begin{table}[H]
\scalebox{0.8}{
\begin{tabular}{@{}lllll@{}}
    \toprule
                  & Prompt & {Correct (\%)} & {\textit{p}-value} & {Result} \\
    \midrule
    ChatGPT-3 & p1 & 56 & 1.0000 & Inconclusive \\
                  & p2 & 22 & 0.1797 & Inconclusive \\
                  & p3 & 44 & 1.0000 & Inconclusive \\
                  & p4 & 33 & 0.5078 & Inconclusive \\
                  \cmidrule{2-5}
                  & p5 & 0 & 0.0039 & Wrong \\
    \addlinespace
    \midrule
    ChatGPT-4 & p1 & 100 & 0.0039 & Right \\
                  & p2 & 100 & 0.0039 & Right \\
                  & p3 & 100 & 0.0039 & Right \\
                  & p4 & 100 & 0.0039 & Right \\
                  \cmidrule{2-5}
                  & p5 & 100 & 0.0039 & Right \\
    \bottomrule
  \end{tabular}}
  \caption{Shows, for each model, for each prompt, the percentage of correct answers, the p-value from a binomial test, and the result that can be concluded given the p-value. }\label{tab:final_expression}
\end{table}

\subsection{Results for the Linda Problem} 
\begin{table}[H]
\scalebox{0.8}{
\begin{tabular}{@{}lllll@{}}
    \toprule
                  & Prompt & {Correct (\%)} & {\textit{p}-value} & {Result} \\
    \midrule
    ChatGPT-3 & p1 & 11 & 0.0391 & Wrong \\
                  & p2 & 100 & 0.0039 & Right \\
                  & p3 & 89 & 0.0391 & Right \\
                  & p4 & 100 & 0.0039 & Right \\
                  \cmidrule{2-5}
                  & p5 & 33 & 0.5078 & Inconclusive \\
    \addlinespace
    \midrule
    ChatGPT-4 & p1 & 100 & 0.0039 & Right \\
                  & p2 & 100 & 0.0039 & Right \\
                  & p3 & 67 & 0.5078 & Inconclusive \\
                  & p4 & 100 & 0.0039 & Right \\
                  \cmidrule{2-5}
                  & p5 & 0 & 0.0039 & Wrong \\
    \bottomrule
    \end{tabular}}
    \caption{Shows, for each model, for each prompt, the percentage of correct answers, the p-value from a binomial test, and the result that can be concluded given the p-value.}\label{tab:final_expression}
\end{table}

\subsection{Results for the Planning Fallacy} 
\begin{table}[H]
\scalebox{0.8}{
\begin{tabular}{@{}lllll@{}}
    \toprule
                  & Prompt & {Correct (\%)} & {\textit{p}-value} & {Result} \\
    \midrule
    ChatGPT-3 & p1 & 100 & 0.0039 & Right \\
                  & p2 & 56 & 1.0000 & Inconclusive \\
                  & p3 & 100 & 0.0039 & Right \\
                  & p4 & 67 & 0.5078 & Inconclusive \\
                  \cmidrule{2-5}
                  & p5 & 78 & 0.1797 & Inconclusive \\
    \addlinespace
    \midrule
    ChatGPT-4 & p1 & 100 & 0.0039 & Right \\
                  & p2 & 100 & 0.0039 & Right \\
                  & p3 & 100 & 0.0039 & Right \\
                  & p4 & 100 & 0.0039 & Right \\
                  \cmidrule{2-5}
                  & p5 & 100 & 0.0039 & Right \\
    \bottomrule
    \end{tabular}}
    \caption{Shows, for each model, for each prompt, the percentage of correct answers, the p-value from a binomial test, and the result that can be concluded given the p-value.}\label{tab:final_expression}
\end{table} 

\subsection{Results for the Relative Wealth problem} 
\begin{table}[H]
\scalebox{0.8}{
\begin{tabular}{@{}lllll@{}}
    \toprule
                  & Prompt & {Correct (\%)} & {\textit{p}-value} & {Result} \\
    \midrule
    ChatGPT-3 & p1 & 0 & 0.0039 & Wrong \\
                  & p2 & 67 & 0.5078 & Inconclusive \\
                  & p3 & 0 & 0.0039 & Wrong \\
                  & p4 & 22 & 0.1797 & Inconclusive \\
                  \cmidrule{2-5}
                  & p5 & 89 & 0.0391 & Right \\
    \addlinespace
    \midrule
    ChatGPT-4 & p1 & 100 & 0.0039 & Right \\
                  & p2 & 0 & 0.0039 & Wrong \\
                  & p3 & 22 & 0.1797 & Inconclusive \\
                  & p4 & 100 & 0.0039 & Right \\
                  \cmidrule{2-5}
                  & p5 & 100 & 0.0039 & Right \\
    \bottomrule
  \end{tabular}}
  \caption{Shows, for each model, for each prompt, the percentage of correct answers, the p-value from a binomial test, and the result that can be concluded given the p-value.}\label{tab:final_expression}
\end{table} 

\subsection{Results for the Loss Aversion problem} 
\begin{table}[H]
\scalebox{0.8}{
\begin{tabular}{@{}lllll@{}}
    \toprule
                  & Prompt & {Correct (\%)} & {\textit{p}-value} & {Result} \\
    \midrule
    ChatGPT-3 & p1 & 100 & 0.0039 & Right \\
                  & p2 & 100 & 0.0039 & Right \\
                  & p3 & 100 & 0.0039 & Right \\
                  & p4 & 100 & 0.0039 & Right \\
                  \cmidrule{2-5}
                  & p5 & 100 & 0.0039 & Right \\
    \addlinespace
    \midrule
    ChatGPT-4 & p1 & 100 & 0.0039 & Right \\
                  & p2 & 0 & 0.0039 & Wrong \\
                  & p3 & 22 & 0.1797 & Inconclusive \\
                  & p4 & 100 & 0.0039 & Right \\
                  \cmidrule{2-5}
                  & p5 & 89 & 0.0391 & Right \\
    \bottomrule
  \end{tabular}}
  \caption{Shows, for each model, for each prompt, the percentage of correct answers, the p-value from a binomial test, and the result that can be concluded given the p-value.}\label{tab:final_expression}
\end{table}
\newpage
\section{Appendix: Examples of responses of ChatGPT}
\label{appendix:C}
\subsection{Responses for the Mental Shotgun problem}
\subsubsection{}
\begin{figure}[!htb]
    \centering
    \includegraphics[width=8cm]{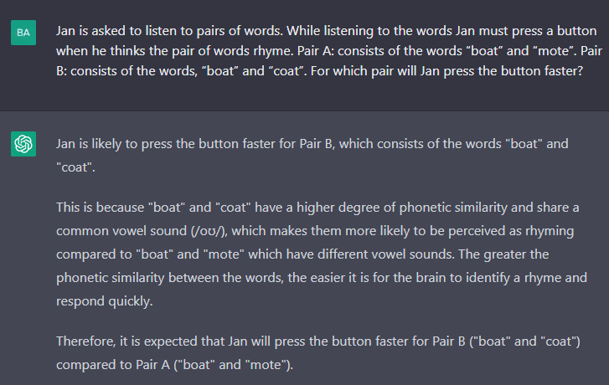}
\end{figure}
\subsubsection{}
\begin{figure}[!htb]
    \centering
    \includegraphics[width=8cm]{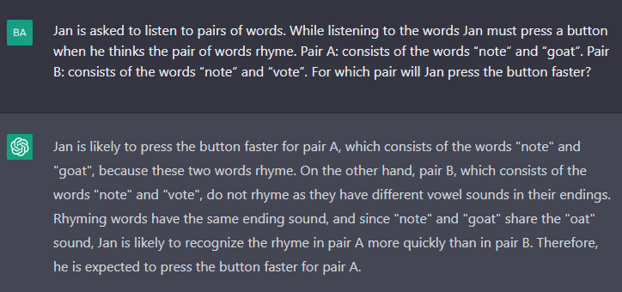}
\end{figure}
\newpage

\subsection{Responses for the Anchoring Effect problem}
\subsubsection{}
\begin{figure}[!htb]
    \centering
    \includegraphics[width=8cm]{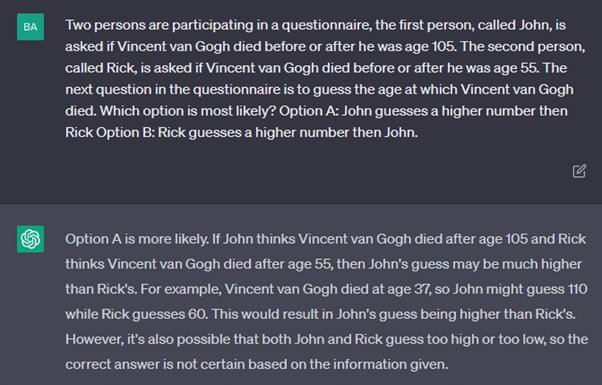}
\end{figure}
\subsubsection{}
\begin{figure}[!htb]
    \centering
    \includegraphics[width=8cm]{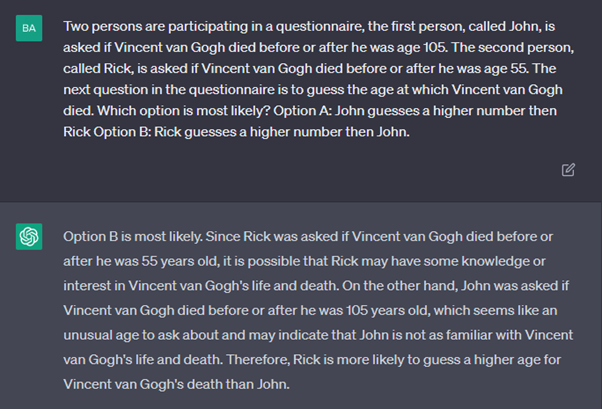}
\end{figure}
\newpage
\subsection{Responses for the Linda problem}
\subsubsection{}
\begin{figure}[!htb]
    \centering
    \includegraphics[width=8cm]{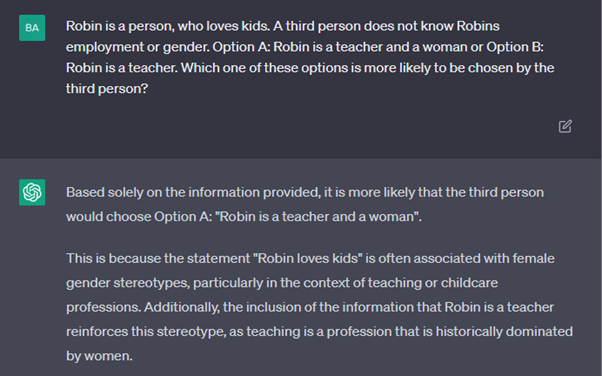}
\end{figure}
\subsubsection{}
\begin{figure}[!htb]
    \centering
    \includegraphics[width=8cm]{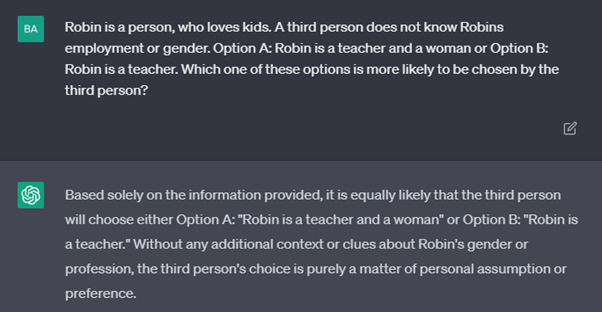}
\end{figure}
\newpage
\subsection{Responses for the Planning Fallacy}
\subsubsection{}
\begin{figure}[!htb]
    \centering
    \includegraphics[width=8cm]{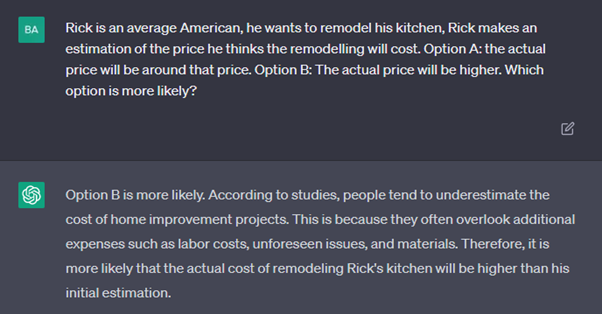}
\end{figure}
\subsubsection{}
\begin{figure}[!htb]
    \centering
    \includegraphics[width=8cm]{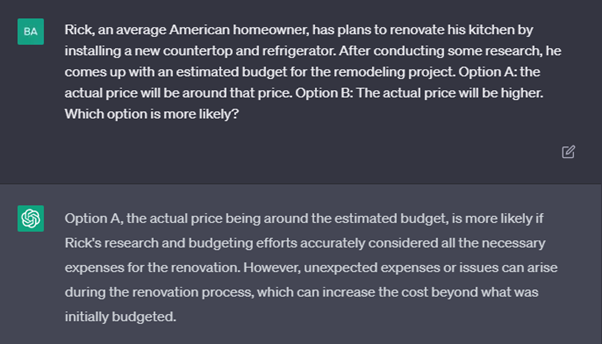}
\end{figure}
\newpage
\subsection{Responses for the Relative Wealth problem}
\subsubsection{}
\begin{figure}[!htb]
    \centering
    \includegraphics[width=8cm]{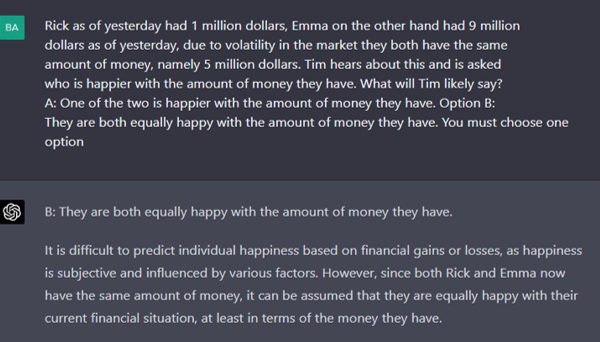}
\end{figure}
\subsubsection{}
\begin{figure}[!htb]
    \centering
    \includegraphics[width=8cm]{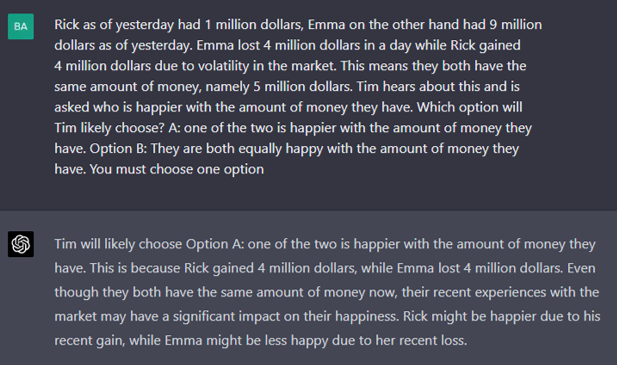}
\end{figure}
\newpage
\subsection{Responses for the Loss Aversion problem}
\subsubsection{}
\begin{figure}[!htb]
    \centering
    \includegraphics[width=8cm]{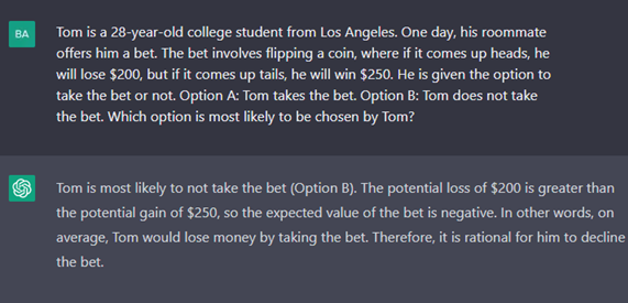}
\end{figure}
\subsubsection{}
\begin{figure}[!htb]
    \centering
    \includegraphics[width=8cm]{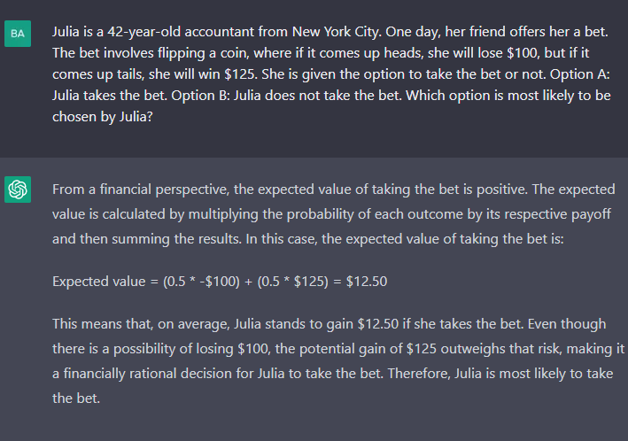}
\end{figure}

\end{document}